\pdfoutput=1

\documentclass[11pt]{article}

\usepackage{ACL2023}

\usepackage{times}
\usepackage{latexsym}

\usepackage[T1]{fontenc}

\usepackage[utf8]{inputenc}

\usepackage{microtype}

\usepackage{inconsolata}

\usepackage{scalerel,graphicx,xparse}
\usepackage{csquotes}
\usepackage{booktabs} 
\usepackage{xspace}
\usepackage{subfig}
\usepackage{xcolor}
\usepackage{CJKutf8}

\newcommand{\chinese}[1]{\begin{CJK*}{UTF8}{gbsn}#1\end{CJK*}}
\newcommand\blfootnote[1]{%
  \begingroup
  \renewcommand\thefootnote{}\footnote{#1}%
  \addtocounter{footnote}{-1}%
  \endgroup
}

\title{
Browsing Lost Unformed Recollections: \\
A Benchmark for Tip-of-the-Tongue Search and Reasoning
}

\author{
Sky CH-Wang$^{\circ}$$^{\bullet}$\quad
Darshan Deshpande$^{\circ}$\quad
Smaranda Muresan$^{\bullet}$\quad
Anand Kannappan$^{\circ}$\quad
Rebecca Qian$^{\circ}$\\
$^{\circ}$Patronus AI\quad
$^{\bullet}$Columbia University\quad \\
\texttt{\{skywang, smara\}@cs.columbia.edu}, \texttt{\{darshan, anand, rebecca\}@patronus.ai},
}

\begin{document}
\maketitle
\begin{abstract}
We introduce \textsc{\textbf{B}rowsing \textbf{L}ost \textbf{U}nformed \textbf{R}ecollections}, a tip-of-the-tongue known-item search and reasoning benchmark for general AI assistants.
BLUR introduces a set of 573 real-world validated questions that demand searching and reasoning across multimodal and multilingual inputs, as well as proficient tool use, in order to excel on.
Humans easily ace these questions (scoring on average 98\%), while the best-performing system
scores around 56\%. 
To facilitate progress toward addressing this challenging and aspirational use case for general AI assistants, we release 350 questions through a public leaderboard, retain the answers to 250 of them, and have the rest as a private test set.\blfootnote{
\href{https://www.huggingface.co/datasets/PatronusAI/BLUR}{https://www.huggingface.co/datasets/PatronusAI/BLUR}
}

\end{abstract}

\section{Introduction}
Have you ever found it difficult to recall the name of something you can vividly describe? Like grasping at the memory of a movie whose title escapes you, even though its fragments remain vividly etched in your memory---the scene of a character gazing into the rain, murmuring something poignant about moments slipping away, or the way neon lights fracture through the downpour, casting the cityscape in wistful, melancholic hues.
Unlike that of traditional search, \textit{tip-of-the-tongue} (ToT) known-item retrieval queries \cite{arguello2021tip} (e.g. Figure \ref{figure1_example}) stand out for their varied information-seeking strategies, requirements of complex multi-hop reasoning, and frequent expressions of uncertainty.
The continued growth of online communities dedicated to posing and resolving such ToT queries, like \texttt{r/TipOfMyTongue}, highlights this ongoing challenge and underscores an aspirational use case for improved solutions \cite{mun2024participAI}.

\begin{figure}[!t]
    \centering
    \begin{displayquote}
    {\footnotesize
    
    \textbf{Query}: I am trying to remember the title of a book I once read. It was published in 2017 and its cover had the image of a snowman looking over some mountains. It had something to do with the search for knowledge. I cannot remember the author of the book but he was a co-author of another book, Conversaciones para Triunfar. What is the title of the book I am looking for?
    
    \textbf{File Input}: None
    
    \textbf{Answer}: Hho the Snowman: The Great Journey
    } \end{displayquote}
    \caption{A sample \textbf{text-only} query and its corresponding answer from our BLUR dataset. An example that is multimodal \textit{on input} is shown in Figure \ref{fig:prompt_response_example}.
    }
    \label{figure1_example}
\end{figure}

In this work, we introduce \textsc{\textbf{B}rowsing \textbf{L}ost \textbf{U}nformed \textbf{R}ecollections} (\textbf{BLUR}),
a multimodal, and multilingual benchmark for general AI assistants featuring 573 curated tip-of-the-tongue questions, validation chains, and corresponding answers. 
These queries represent real-world information needs, are easy to create and validate, and assess the capabilities of systems to reason under uncertainty.
To excel on this benchmark, a general AI assistant will need to excel at multi-hop reasoning over multimodal information, scope potential uncertainty regions associated with questions, and be proficient in general tool use.
For system designers, doing so will signify meaningful advancements in these capabilities \cite{wagstaff2012machine} as well as represent a major step towards addressing this challenging and aspirational real-world use case \cite{liu-etal-2024-ecbd}.
Human participants consistently score near-perfect results (98\%) on these questions, while the best-performing system tested scores around 56\%. Surprisingly, we find that the best agentic systems performed only marginally better than the base language models that power them, suggesting that large gaps remain in effective reasoning around tool use for this task.
We open source a developer subset of 350 annotated questions from BLUR---while maintaining both public and private test sets---to avoid contamination and provide a fair and unbiased evaluation.

\section{Related Work}

\paragraph{Known-Item Retrieval.} In information retrieval, tip-of-the-tongue known item retrieval has emerged in recent years as a new 
challenge task for information retrieval systems. Using online community-based posts and answers from sources like Reddit \cite{bhargav2022s}, datasets for known item retrieval have mostly centered around movie \cite{arguello2021tip}, music \cite{bhargav2023music}, and book \cite{lin-etal-2023-decomposing} identification.
Qualitative coding schemes and analyses of posts in these communities reveal that people mostly use such platforms for general entertainment or leisure information identification \cite{meier2021towards,frobe2023large,bogers2024understanding}. 
Using these datasets, recent work has explored leveraging LLMs to break down complex queries into their essential components through prompting \cite{lin-etal-2023-decomposing}, within content re-ranking \cite{borges2024generalizable}, 
and in generating clarification questions \cite{Chi2024CLARINETAL}. Furthermore, \citet{kim-etal-2024-intended} illustrate the potential for advancements in ToT identification to support patients with anomia, a difficulty in identifying the names of items. 

Despite this wealth of online data, directly translating these sources into a task that both validly measures the ability of AI assistants to perform this task and evaluates underlying system capabilities remains a challenge. 
The first challenge lies in answer ambiguity. Multiple items may accurately match a user's query, as evidenced by the variety of responses typical posts receive. Measuring \textit{correctness} becomes difficult 
with query underspecification, causing the number of \textit{valid} answers to be unknown in a near-infinite set of possibilities \cite{arguello2021tip}.
The second challenge lies in the valid measurement of underlying system capabilities. Answers are often provided without explanations of their derivation, making it unclear which abilities a system must possess to perform well and obscures the skills ultimately being tested for a benchmark dataset \cite{liu-etal-2024-ecbd}.
The third challenge lies in data contamination. As most LLMs that power these systems are trained on publicly available internet data, contamination risks compromise the validity of evaluation results \cite{sainz-etal-2023-nlp}. 
Here, we tackle these three challenges by ensuring questions have \textit{unambiguous answers}, incorporate human reasoning chains for more robust \textit{capability grounding}, and keep a private test set to \textit{avoid data contamination}.

\paragraph{Multi-Hop Reasoning and Retrieval.} Complex real-world tasks often demand more than single-step retrieval; they require sophisticated, multi-hop reasoning that aggregates information from multiple sources and modalities. ToT exemplifies this complexity, calling for iterative data gathering, reasoning, and comparative analysis~\cite{arguello2021tip}. While existing benchmarks highlight the importance of multi-step reasoning for tasks such as coding, data analysis, and open-ended question answering \cite{petroni-etal-2021-kilt, jimenez2024swebench, hu2024infiagent}, they do not fully capture how LLMs can dynamically plan and retrieve content from multimodal tools, including map or video platforms. \citet{liu2024agentbench}, \citet{yoran2024assistantbench}, and \citet{NEURIPS2023_mind2web} extend benchmarking by introducing operating system access, map navigation, and other utilities, thereby underscoring the significance of multimodal reasoning. However, these approaches rarely address ambiguous queries, misleading descriptions, or multi-domain exploration akin to real-world conditions. BLUR fills this gap by requiring complex, scenario-based reasoning and multilingual capabilities over a broad spectrum of multimodal tools.

\paragraph{Evaluation.} 

Large language models are saturating measurements of their capabilities at an increasingly unprecedented rate \cite{kiela2023plottingprogress}, spurring a surge of techniques aimed at pinpointing their vulnerabilities. Methods like adversarial and dynamic benchmarking rely on iterative, model-in-the-loop data collection to produce increasingly difficult tasks for existing models \cite{kiela-etal-2021-dynabench, wei2024measuring}. While these approaches successfully expose weaknesses, they often generate tasks that lack ecological validity \cite{bowman-dahl-2021-will, phang-etal-2022-adversarially, de2020towards, liu-etal-2024-ecbd} and do not reflect how models will be used in real-world scenarios \cite{liao2023rethinking}. Our field of evaluation must move beyond merely identifying shortcomings and instead broaden our evaluation strategies to address \textit{genuine} cross-capability performance in \textit{real-world contexts}~\cite{bowman-dahl-2021-will}.

Because human evaluation is expensive and hard to scale, many assessments resort to automated or simplified methods, including model-based judgments \cite{zheng2023judging} or string matching and multiple-choice questions for correctness checks. However, multiple-choice formats can obscure flawed reasoning, since a model may arrive at the correct answer without demonstrating a sound solution path \cite{mialon2024gaia}.
Although substantial effort has gone into evaluating the specialized abilities of LLMs \cite{hendrycks2021measuring, rein2024gpqa}, recent work has shifted attention to measuring general-purpose proficiency. Most of these efforts rely on closed systems or predefined API interactions \cite{patil2023gorilla, li-etal-2023-api, liu2024agentbench}, which may reward familiarity with specific APIs over broader, more transferrable competencies. To address these limitations, recent research---such as OpenAGI \cite{ge2024openagi} and GAIA \cite{mialon2024gaia}, along with tasks like OSWorld \cite{OSWorld} and systems like Claude 3.5 Sonnet’s computer use \cite{claude_computer_use}---advocates broader evaluations by foregoing strict API constraints and embracing more open real-world interactions. Similarly, Dynasaur \cite{nguyen2024dynasaur} investigates the on-the-fly creation and execution of task-specific tools, employing code as a unifying representation.
In BLUR, we focus on \textit{human-driven tasks}, prioritize simplified automated evaluation, and center on aspirational general AI assistant capabilities.

\begin{figure*}[!t]

    \centering
    \begin{tabular}{ccc}
    \subfloat[Prompt Domains]{\includegraphics[width=0.30\textwidth]{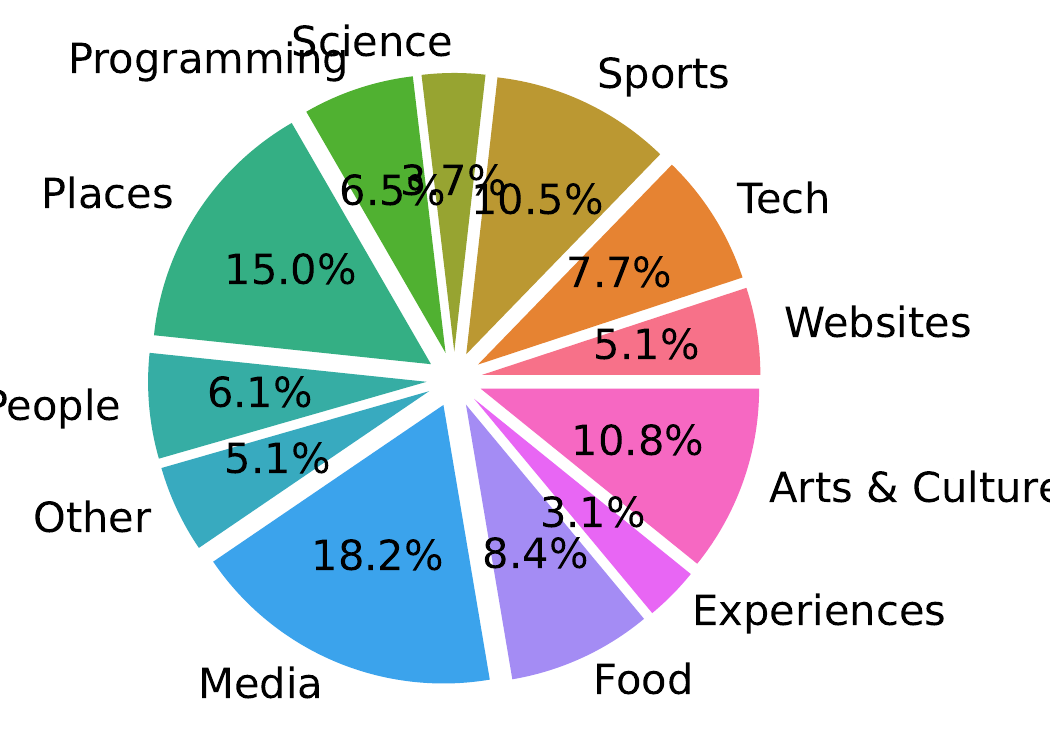}} &  
    \subfloat[File Extensions]{\includegraphics[width=0.30\textwidth]{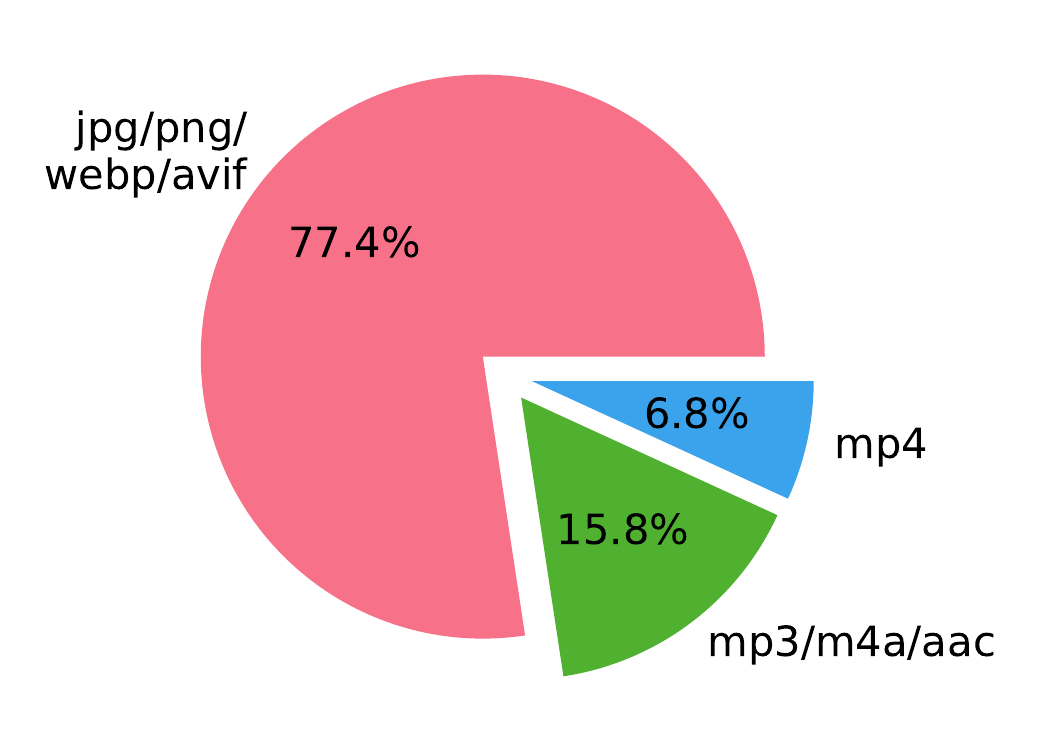}} &    
    \subfloat[Non-English Languages
    ]{\includegraphics[width=0.30\textwidth]{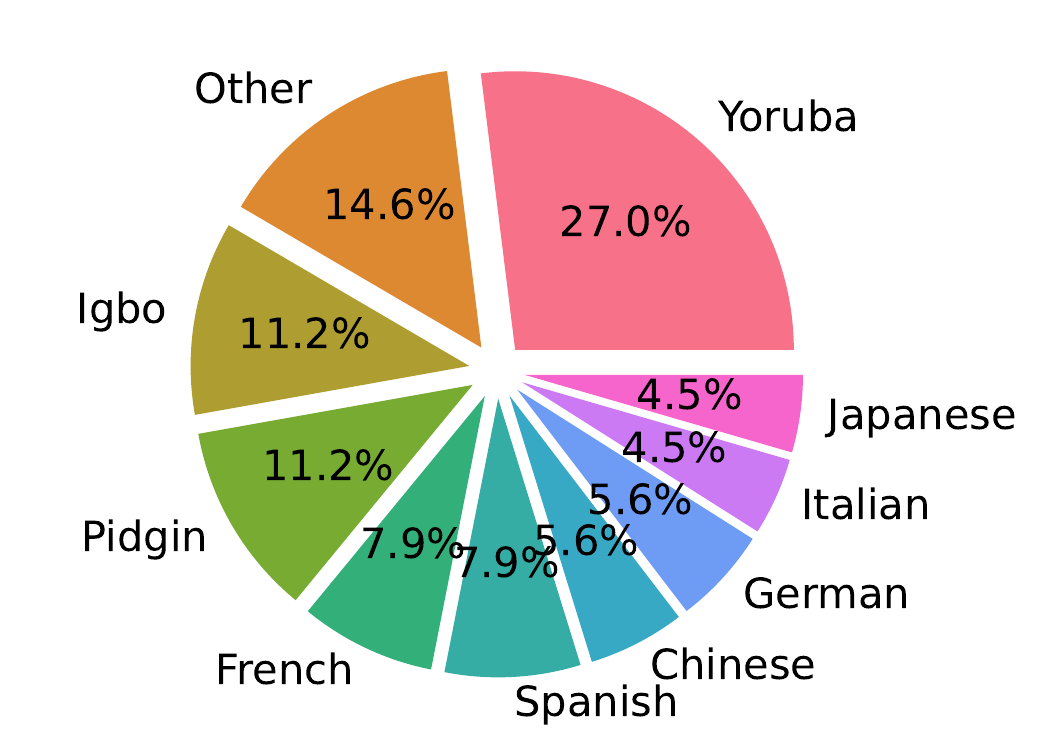}}
    \end{tabular}

    \caption{Distribution of prompts across \textbf{topical domains}, \textbf{file extensions} for prompts with attached files, and non-English-language-only prompt \textbf{languages}. Note that 75\% of prompts are text-only without attached files, and 70\% are exclusively in English.}
    \label{domains_extensions_languages}
    
\end{figure*}

\section{The BLUR Benchmark}

BLUR is a benchmark designed to evaluate general AI assistants on a series of tip-of-the-tongue known-item retrieval tasks. It comprises 573 text-based questions covering a variety of topical domains, with 25\% including additional input from a different modality (e.g., image, audio, or video) in the form of a file attachment. Although all prompt instructions are primarily in English, 30\% of the queries are multilingual in the sense that they either directly feature descriptions written in other languages or reference items that are primarily available online in a different language other than English. Answers to these queries are validated to be unambiguously correct and formatted as concise strings for automatic verification either through a direct weak string match or with the help of an LLM-Judge. 
BLUR is easy to use, requiring only a zero-shot prompt with provided scaffolding against an AI assistant for answer elicitation. Success on BLUR demands strong multi-hop reasoning, effective tool use, and prompt uncertainty and ambiguity handling.
Figure \ref{domains_extensions_languages} illustrates the distribution of prompts across domains, file extensions, and languages.
The following sections outline the design principles, question guidelines, dataset composition, and evaluation methodology.
The set of guidelines given to annotators and annotator details are shown in Appendix Section \ref{sec:annotation_details}.

\subsection{Goals and Design Choices}

To construct this dataset, we invited annotators to reflect on recent or current instances where they struggled to recall the name of something. Annotators were asked to describe everything they could recall about the item in question, framing it as a prompt they might use to seek help online or from a friend. Annotators were also given the option to upload a file in addition to providing a text input if they wished.
We then tasked the writer with locating the item whose name they struggled to recall. Separately, a different annotator (the validator) was challenged to identify the item based solely on the original query provided by the writer. 
Both annotators were given access to a web browser and documented their search process step by step.
If the validator's answer matched the writer's, the prompt was included in the final dataset. Otherwise, we presented the validator with the correct answer along with the writer's search steps and evaluated \textit{posthoc agreement}—whether the validator acknowledged their error and could clearly articulate their mistake. If posthoc agreement was achieved, the prompt was included in the final dataset; otherwise, it was discarded. Prompts were finally minimally edited to standardize formatting and correct typos. 

\paragraph{Unambiguous Answers.} The majority of the effort in this two-stage dataset creation process---prompt writing followed by validation---focused on ensuring that the prompts were unambiguous, meaning they led to a single, correct answer \cite{rein2024gpqa, mialon2024gaia}. In doing so, we deliberately avoided adversarial dataset construction, as it not only obscures the specific abilities benchmarks aim to measure \cite{bowman-dahl-2021-will}, but also undermines ecological validity \cite{de2020towards}.
While the uniqueness of these items is sure to evolve over time---for example, a building might be replaced or new film releases might echo details in earlier films---we address potential \textit{eventual ambiguities} by specifying the exact date the question was asked as part of the evaluation prompt scaffold (Figure \ref{fig:prompt_response_example}).
In addition, we periodically re-examine the \textit{sources of truth} that validators relied upon to confirm their answers, ensuring that these details remain accessible and verifiable by systems. 

\paragraph{Multimodal and Multilingual.} While annotators were instructed to write their prompts in English, no language restrictions were placed on the details of the items remembered. Approximately 30\% of our dataset is notably multilingual. This includes cases where descriptions are written in other languages or where the descriptions are in one language but the item itself belongs to a different language. 
Similarly, 25\% of our dataset is multimodal \textit{at input}, featuring prompts accompanied by file attachments rather than being exclusively text-based. This contrasts with Reddit-based ToT datasets in information retrieval \cite{meier2021towards}, where attachments---though originally included in the posts by users---are typically removed or ignored during preprocessing. Files included sketches of the items recalled, similar images found online, video and audio files in which the item appeared in, and more. 
Note that, in addition to the explicit multimodal understanding required to process file inputs, a majority of queries in BLUR also require reasoning over multimodal sources of information encountered in web searches (images, videos, and more) despite being only text-based. 

\paragraph{Ease of Use.} 
Answers in the dataset are concise, consisting of a single string that can be evaluated for correctness using a weak string match with a LLM Judge (similar to \citet{wei2024measuring}). Prompts are designed for zero-shot answering and evaluation, structured within a question scaffold (Figure \ref{fig:prompt_response_example}) that constrains the date the question was asked as well as the answer format. 
In practice, when faced with a ToT problem, user preferences may intuitively be split between favoring single-turn interactions or iterative engagement with a general AI assistant, correcting its mistakes over multiple conversation rounds to arrive at the correct answer. Here, we focus on the former usage scenario, curating questions depicting real information needs that \textit{can} be resolved within a single turn.

\begin{figure}
    \centering
    \includegraphics[width=0.98\linewidth]{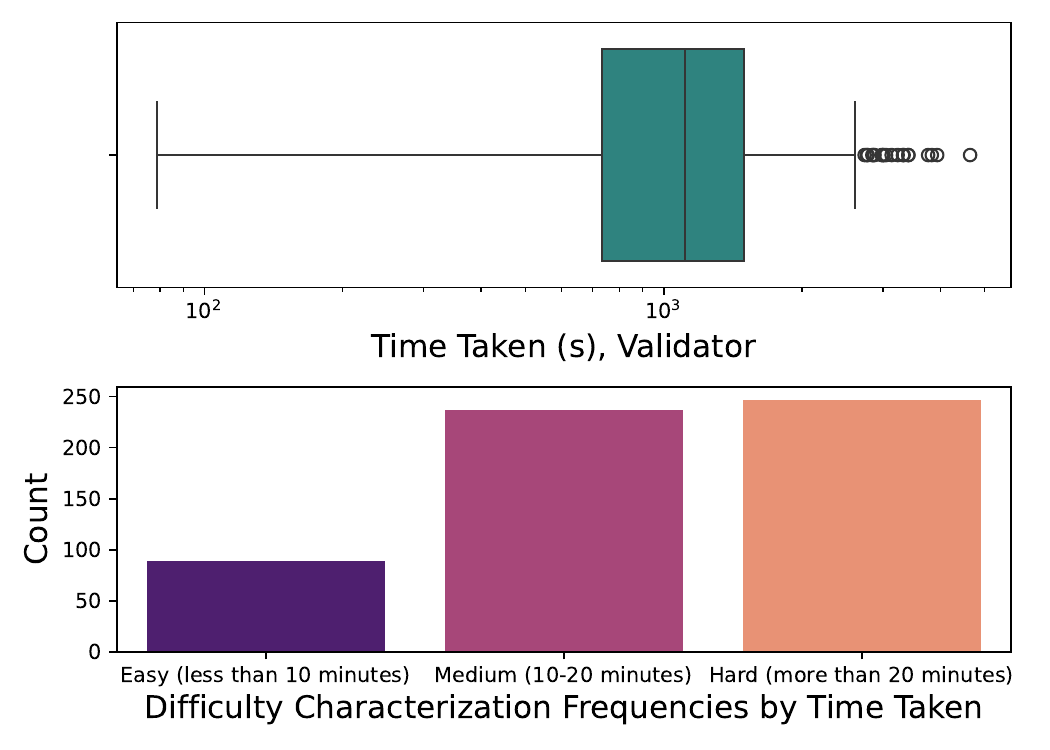}
    \caption{Distribution of \textbf{response times} taken by validators to answer questions (top) and the breakdown of \textbf{query difficulty} levels (bottom) as characterized by splits based on those times.}
    \label{fig:difficulty_splits}
\end{figure}

\begin{figure*}[!ht]
    \centering

    \begin{displayquote}{\footnotesize
    \textbf{Prompt Scaffold}: You are a general AI assistant assigned to help a user identify an entity they have forgotten. Below is the user’s query, asked on December 15th 2024, consisting of a textual description possibly accompanied by an optional file attachment. Approach the task methodically, utilizing all available tools, and document each step you take in detail. Conclude your response with the following format: Final Answer: [Your final answer]. Your final answer should accurately provide the answer to the user's query as a short string.
    
    \textbf{Query}: I visited a bank in Ibadan with a friend, but I can’t recall its name or location. It was my first time in Ibadan, so my memory of the place is a bit vague. However, I remember taking a picture of an attractive building located opposite the bank. I’ll attach the picture, can you help me identify the bank and its address?
    
    \textbf{File Input}: } \end{displayquote}

    \includegraphics[width=0.35\textwidth]{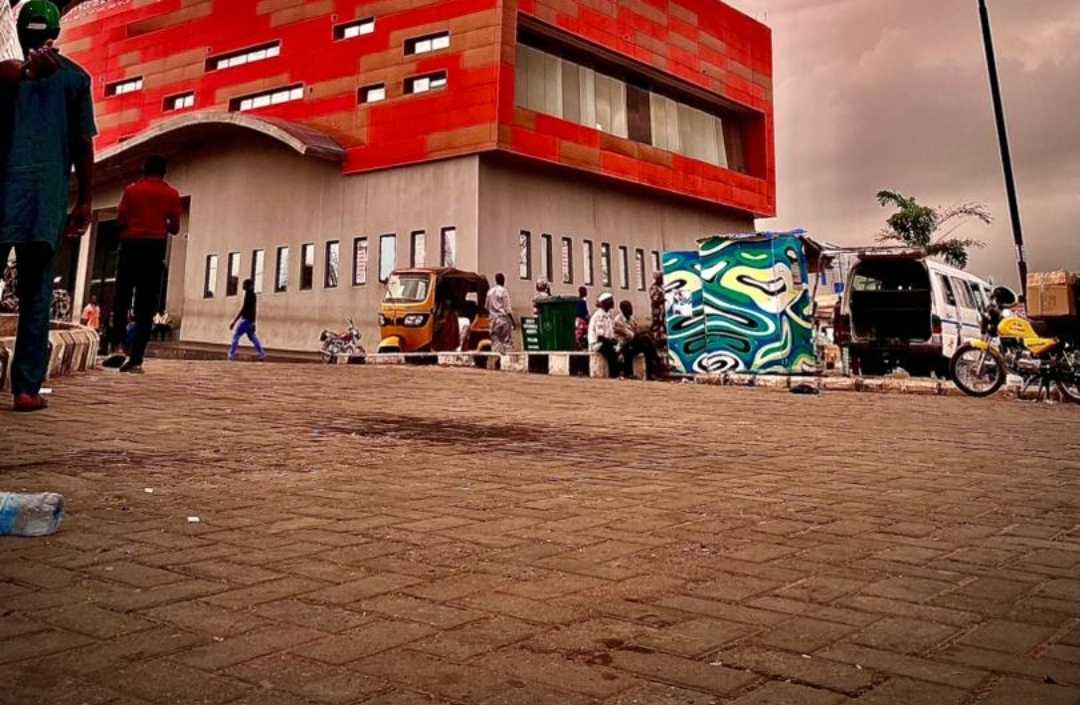}

    \begin{displayquote}{\footnotesize
    \textbf{ChatGPT 4o}: \colorbox{gray}{from PIL import Image...}

    [Matplotlib Chart]

    \colorbox{orange}{Searching Google Maps...}

    Based on the information provided and the image analysis, it appears that the bank you visited in Ibadan is the First Bank - Ibadan Ring Road Branch, located at 15 MKO Abiola Way, Oluyole, Ibadan 200212, Oyo, Nigeria. [Business List]

    \textbf{Final Answer}: First Bank - Ibadan Ring Road Branch, 15 MKO Abiola Way, Oluyole, Ibadan 200212, Oyo, Nigeria

    \textbf{Ground Truth}: Zenith Bank: 108, Lagos bye-pass Challenge, Ibadan.

    \textbf{LLM Judge}: Weak String Match Result: \colorbox{red}{FALSE}
    
    } \end{displayquote}
    
    \caption{\textbf{Evaluation instance example} of a medium-difficulty query with a file input. The prompt scaffold is shown, along with the output provided by ChatGPT 4o and LLM Judge weak string match result. The step-by-step validation chain that a human validator took to answer this query is shown in Figure \ref{fig:validation_chain}. Code execution in output is highlighted in \colorbox{gray}{gray}, while web search is highlighted in \colorbox{orange}{orange}; both are truncated for simplicity.}
    \label{fig:prompt_response_example}
\end{figure*}

\begin{figure}[!ht]
    \centering
    \begin{displayquote}{\footnotesize

    1. I upload[ed] the picture on Google Lens to search.
    
    2. I look[ed] through the results and click[ed] on \href{https://www.youtube.com/watch?v=pkIEisnRsqA}{this} YouTube video.
    
    3. I watch[ed] the video and compare[d] it to the uploaded image. I confirmed the location is Challenge Bus Terminal.
    
    4. I searched for ``Challenge Bus Terminal'' on Google Maps.
    
    5. I looked at \href{https://www.google.com/maps/place/Challenge+Bus+Terminal/@7.3460999,3.8769063,18.29z/data=!4m6!3m5!1s0x10398d9057ac73fb:0xf571e27835f651a8!8m2!3d7.3452289!4d3.8780037!16s%2Fg%2F11r_88h2bp?entry=ttu&g_ep=EgoyMDI0MTIwNC4wIKXMDSoASAFQAw%3D%3D}{the results} and compared them with the uploaded image. I confirmed [that] they match. 
    
    6. I \href{https://www.google.com/maps/place/Challenge+Bus+Terminal/@7.3457408,3.8775057,3a,28.6y,305.81h,93.74t/data=!3m7!1e1!3m5!1sEHiVi9gzJuZRTAANKQynKQ!2e0!6shttps:%2F%2Fstreetviewpixels-pa.googleapis.com%2Fv1%2Fthumbnail%3Fcb_client%3Dmaps_sv.tactile%26w%3D900%26h%3D600%26pitch%3D-3.739911644974484%26panoid%3DEHiVi9gzJuZRTAANKQynKQ%26yaw%3D305.8090251388961!7i13312!8i6656!4m14!1m7!3m6!1s0x10398d9057ac73fb:0xf571e27835f651a8!2sChallenge+Bus+Terminal!8m2!3d7.3452289!4d3.8780037!16s%2Fg%2F11r_88h2bp!3m5!1s0x10398d9057ac73fb:0xf571e27835f651a8!8m2!3d7.3452289!4d3.8780037!16s%2Fg%2F11r_88h2bp?entry=ttu&g_ep=EgoyMDI0MTIwNC4wIKXMDSoASAFQAw%3D%3D}{confirm[ed] through [Google] [S]treet [V]iew} that there's a Zenith Bank opposite the terminal.
    
    7. I [G]oogled ``Zenith bank challenge Ibadan address[.]''
    
    8. I looked through the results and clicked on \href{https://www.zenithbank.com/branches/#:~:text=108%2C%20Lagos%20bye%2Dpass%20Challenge%2C%20Ibadan.}{this} link.
    
    9. I looked through the website and confirmed [that] the address is 108, Lagos bye-pass Challenge, Ibadan.
    
    10. I \colorbox{green}{confirm} [that] the answer is Zenith Bank, 108, Lagos bye-pass Challenge, Ibadan.
    
    } \end{displayquote}
    \caption{Steps that a human validator took to answer the query shown in Figure \ref{fig:prompt_response_example}. The time spent by the validator to arrive at the answer was fifteen minutes and forty-five seconds. 
    }
    \label{fig:validation_chain}
\end{figure}

\paragraph{Hard to Game.}
Data contamination and memorization pose significant challenges to accurately assessing a system's capabilities \cite{sainz-etal-2023-nlp}. 
While it is not entirely impossible that similar questions or answers to some questions may exist in pre-training data by chance, the possibility of checking reasoning traces provides an additional mechanism to assess tool use and detect memorization behavior. 
Finally, we offer multiple leaderboard variations: one with a public validation set (including answers), another with a public test set (excluding answers), and a fully private test set (with both questions and answers withheld). 
These mitigations are designed to help ensure that improvements in system performance on this benchmark dataset reflect genuine advancements in system capabilities.
Should memorization still occur despite these safeguards, practitioners can easily create new questions by following the guidelines outlined in Appendix \ref{sec:annotation_details}.

\subsection{Composition}

Figure \ref{domains_extensions_languages} illustrates the distribution of queries across topical domains, as categorized by the authors using a grounded coding scheme. It also highlights the file extensions associated with queries that include file inputs and the language capabilities identified by annotators as necessary to answer each query accurately.

\paragraph{Query Domains and Capabilities.} Consistent with prior studies of posts in ToT information communities online \cite{frobe2023large}, most prompts fall into the \textit{Media} category, with queries about songs, movies, videos, shows, and books, among others. 
We characterize the capabilities required of general AI systems to perform perfectly on BLUR via grounded coding on the validation chains annotators recorded when answering queries. Below, we report the capabilities, as well as specify concrete examples of tools used by annotators. Note that tools may fall into more than one capability characterization. 

\begin{table*}[!ht]
    \centering
    \begin{tabular}{ l | c c | c c c | c }
        \toprule
        \multicolumn{7}{c}{\textbf{BLUR Performance ($\uparrow$)}} \\
        \midrule
        Model / System & $Q_{T}$ & $Q_{F}$ & $H_{E}$ & $H_{M}$ & $H_{H}$ & Overall \\
        \midrule 
        Llama-3.1-405B & 0.34 & 0.17$^{\circ}$ & 0.35 & 0.32 & 0.25 & 0.30 \\
        claude-3-5-sonnet-20241022 & 0.44 & 0.28$^{\bullet}$ & 0.42 & 0.42 & 0.36 & 0.40 \\
        gpt-4o-2024-11-20 & 0.42 & 0.28$^{\bullet}$ & 0.39 & 0.43 & 0.35 & 0.38 \\
        o1-2024-12-17 & 0.54 & 0.36$^{\bullet}$ & 0.56 & 0.52 & 0.44 & 0.49 \\
        DeepSeek-R1 & 0.45 & 0.27$^{\circ}$ & 0.46 & 0.44 & 0.35 & 0.41 \\
        \midrule 
        Microsoft Copilot & 0.29 & 0.23$^{\bullet}$ & 0.29 & 0.32 & 0.22 & 0.27 \\
        Mistral Le Chat & 0.40 & 0.27$^{\bullet}$ & 0.47 & 0.38 & 0.32 & 0.37 \\
        Perplexity Pro Search & 0.31 & 0.15$^{\bullet}$ & 0.29 & 0.29 & 0.24 & 0.27 \\
        ChatGPT-4o & 0.53 & 0.36\hphantom{$^{\bullet}$} & 0.60 & 0.52 & 0.41 & 0.49 \\
        \midrule
        HuggingFace Agents + Claude 3.5 Sonnet & 0.61 & 0.41$^{\bullet}$ & 0.60 & 0.56 & 0.54 & 0.56 \\
        DynaSaur + GPT-4o & 0.58 & 0.27\hphantom{$^{\bullet}$} & 0.61 & 0.52 & 0.44 & 0.50 \\
        Operator & 0.57 & 0.46$^{\bullet}$ & 0.56 & 0.56 & 0.52 & 0.54 \\
        \midrule
        Search Engine & 0.05 & 0.03$^{\bullet}$ & 0.08 & 0.05 & 0.02 & 0.04 \\
        \textbf{Human} & \textbf{0.98} & \textbf{1.00}\hphantom{$^{\bullet}$} & 0.98 & 0.98 & 0.99 & \textbf{0.98} \\
        \bottomrule
    \end{tabular}
    \caption{
        \textbf{System and model performance} on the BLUR benchmark. $Q_{T}$ and $Q_{F}$ denote performance on text-only queries and queries with file inputs, respectively. System support for file inputs is indicated, where $\circ$ signifies that the system does not support file uploads and $\bullet$ denotes partial support of certain file type extensions; the absence of a circle denotes that all file type uploads are supported. $H_{E}$, $H_{M}$, and $H_{H}$ represent system performance on \textit{easy}, \textit{medium}, and \textit{hard} query difficulty subsets, respectively. 
    }
    \label{tab:main_results}
\end{table*}

\begin{itemize}
    \item \textbf{Web Browsing}: Internet-based browsing and navigation. Examples include Google Search, Google Maps, and Wikipedia. 
    \item \textbf{Multimodality}: Understanding visual and audio information. Examples include Youtube, Spotify, Bing images, Optical Character Recognition, Google Street View, Reverse Image Search, Google Lens, and Speech to Text. 
    \item \textbf{Multilinguality}: Comprehension of textual information in languages other than English. Examples include Google Translate, Microsoft Translator, and webpage translation.
    \item \textbf{File Reading}: Understanding information present in files of various types. Examples include VLC, iTunes, and Photo Viewer.
\end{itemize}

\paragraph{Difficulty.} The time validators took to answer these queries naturally serves as a proxy of their difficulty level for humans.
Based on these times, we divided the dataset into three difficulty levels: \textit{easy}, for questions resolved in under 10 minutes, \textit{medium}, for those requiring 10 to 20 minutes, and \textit{hard}, for those taking over 20 minutes to answer. Figure \ref{fig:difficulty_splits} illustrates the distribution of times taken by annotators to arrive at their answers, along with the number of queries in each difficulty category.

\section{Evaluation}

\paragraph{Setting.} Answers to queries take the form of short strings. For consistency across evaluations, we provide a zero-shot prompt scaffold that is used to inform systems about the format of the answer output (shown in Figure \ref{fig:prompt_response_example}) and the date that the question was asked. Given a returned answer and the correct ground truth answer, a weak string match is performed to check for correctness, which takes the form of a Llama 3.2-based LLM Judge within a scoring function, implemented alongside the leaderboard.
The prompt for this judge is shown in Appendix Section \ref{sec:prompts}; human validation confirms the accuracy of this judge to be accurate on average 98\% of the time.

Using the prompt scaffold shown in Figure \ref{fig:prompt_response_example}---unless systems have their own prompt scaffold to return a single, final answer---we evaluate
ChatGPT-4o \cite{gpt4o}, Perplexity Pro \cite{perplexitypro}, Mistral Le Chat \cite{mistrallechat}, Microsoft Copilot,\footnote{Accessed through \href{https://copilot.microsoft.com/}{https://copilot.microsoft.com/}}
HuggingFace Agents with Claude 3.5 Sonnet \cite{hfagents},\footnote{Specifically, the implementation that achieves the top attempt on the GAIA leaderboard as described in {https://huggingface.co/blog/beating-gaia}} DynaSaur with GPT-4o \cite{nguyen2024dynasaur}, and Operator \cite{cua2025},
each with their default settings and sets of tools. We compare the performance of these systems against human validator performance, base LLM performance with Llama 3.1 405B \cite{grattafiori2024llama}, GPT-4o and o1 \cite{gpto1}, Claude 3.5 Sonnet \cite{claude_computer_use}, DeepSeek r1 \cite{guo2025deepseek}, and web search, where we type the questions onto a webpage and assess if the correct answer may be obtained from the first page of displayed results. For robustness, due to inherent system stochasticity, we run each system on every question three times and report the average of the results.
We assess our human baseline by measuring the rate at which validators reach the correct answer on their first attempt.

\begin{figure*}[!ht]
    \centering
    \includegraphics[width=0.98\linewidth]{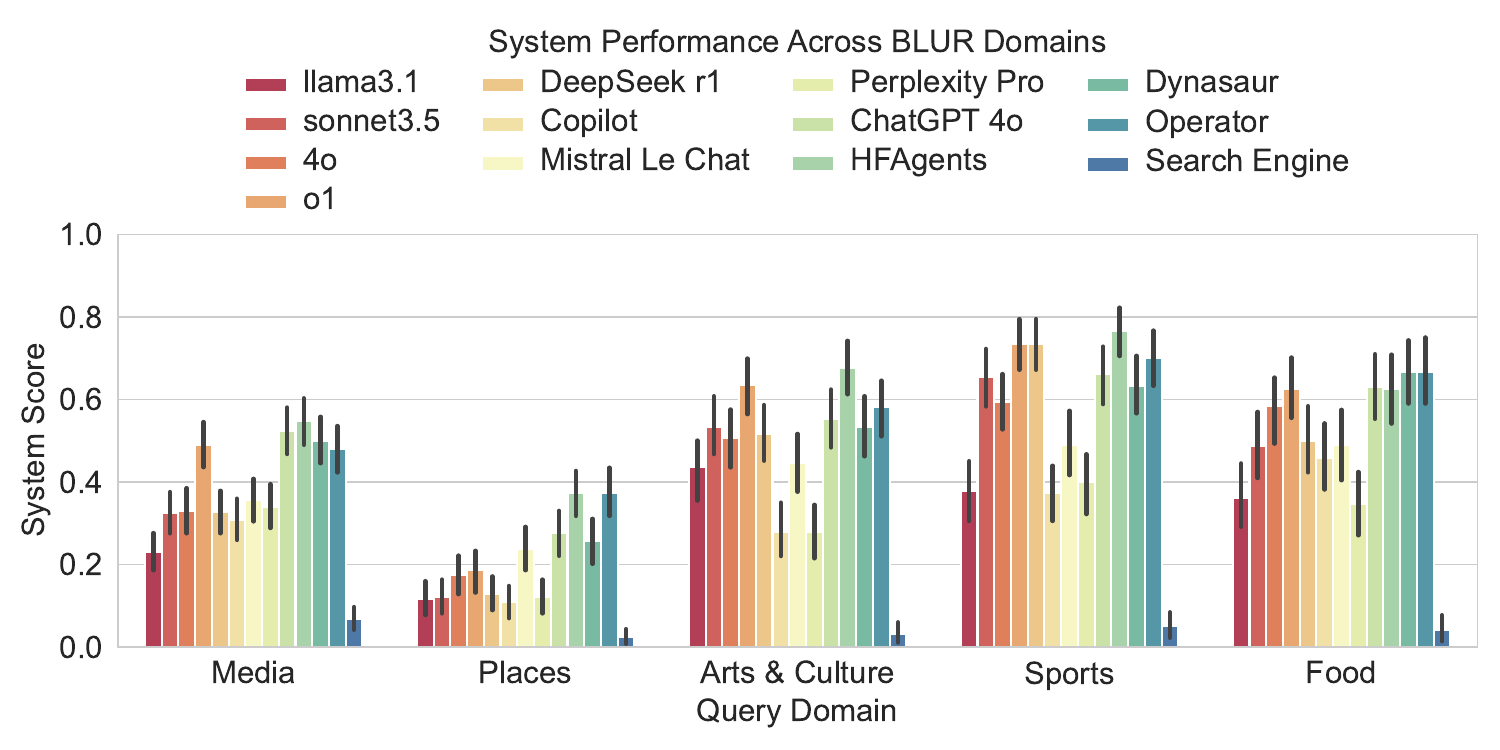}
    \caption{\textbf{System performance} as stratified across the five most frequent \textbf{domain categories}. 95\% confidence intervals are shown. Queries about \textit{places} are less likely to be captured by the typical parametric knowledge derived from internet text, which means they often require proficient tool usage and are consequently more challenging for current systems to answer accurately. }
    \label{fig:performance_domain}
\end{figure*}

\paragraph{Results.} 
As shown in Table \ref{tab:main_results}, overall, peak system performance reaches approximately half of human performance, with consistent declines across higher difficulty tiers (Figure \ref{fig:difficulty_splits}). Questions that take humans longer to solve are, on average, more often incorrectly answered by existing agent systems.
Notably, systems with tool use enabled show only marginal (+7\%) improvements over the strongest base models that lack tool use capabilities. 
Across the most frequent query domains (Figure \ref{fig:performance_domain}), queries around \textit{places} stand out as the hardest for existing systems and models to answer correctly, while queries about \textit{sports} and \textit{food} are comparatively easier. 
Together, these observations point to the following conclusions.
One, \textit{existing forms of parametric knowledge alone prove effective at answering real-world tip-of-the-tongue queries in \textbf{certain} domains:}
Although the exact prompts for queries do not appear online in textual form, the underlying details of their queried items often do. With a reasoning chain for DeepSeek r1 shown in Appendix Section \ref{sec:reasoning_traces}, our findings demonstrate that the most successful models \textit{can} assemble fragments of their learned parametric knowledge to reach correct answers for various query types that have this characteristic. Despite this, parametric knowledge alone is insufficient, particularly when dealing with scenarios that require information either distinct from web-based training data or more current than the model's training cutoff date.
Two, \textit{models fail to reason effectively about which tools to employ and how to employ them.} This behavior along with ineffective comprehension of information returned by tools in the context of given queries, results in performance that is only marginally better than if they used no tools at all. As observed in Appendix \ref{sec:reasoning_traces}, Operator misunderstands the initial search result for travel sites launched in 2000s and continues to pursue an incorrect result for Expedia.
Overall, these results illustrate that though current models are effective at answering certain types of tip-of-the-tongue queries, they still greatly underperform---particularly in comparison to humans---when faced with the full breadth and complexity of this challenging real-world information need.
In the following section, we note qualitative insights into system failures and examine key challenges.

\section{Discussion}

\paragraph{Where do systems fail?} A qualitative analysis of reasoning logs reveals several areas of failure in current systems capable of tool use. 
The first area lies in \textit{contextual understanding}. Even with multimodal capabilities, systems can misinterpret or overlook critical details. For instance, in searching for a ``grey-colored battery pack with a black top,'' Operator incorrectly identified a grey battery pack with \textit{black buttons} as that item—an error that indicates lingering gaps in contextual comprehension and can propagate to downstream answers.
The second area lies in \textit{orchestration}. Some systems prematurely terminate their search after finding an item that satisfies only part of the specified constraints. In contrast, systems like HuggingFace Agents iteratively verify whether additional constraints can be met before returning a final answer. Perplexity Pro, by comparison, performs a single retrieval-augmented generation step; if the correct answer does not appear in the retrieved documents, the system simply states it cannot answer the query. Their relative performance differences highlight the value of effective orchestration for query attempts and validation in succeeding at this task.
The third area lies in dealing with \textit{tool failures}. Tools can fail---on occasion, API rate limits or access restrictions (e.g., YouTube sometimes being inaccessible to Operator) can block a system from retrieving key information. Instead of seeking alternate sources, systems were often observed to become stuck in a loop of repeated attempts to access the same site, ultimately failing to retrieve the needed data.
Finally, the fourth area lies in dealing with \textit{long contexts}. Many queries require aggregating and verifying information from multiple sources, leading to lengthy action sequences. Over time, systems may lose track of the original query or become ensnared in an infinite loop of searches, preventing them from reaching a correct conclusion. 
These observations highlight existing areas of improvement for models to improve further at this challenging task. 

\paragraph{Is Tool Use \textit{Really} Needed?} 
Although the best models without tool-use capabilities (i.e. o1) often succeeded by effectively leveraging their existing parametric knowledge to piece together the answers to queries, their failures in domains like \textit{places} reveal the limits of relying solely on fixed pretraining knowledge. One alternative to further push the performance of these language-only models is to \textit{augment pretraining data} even further with existing tools; i.e., through techniques like generating multiple textual descriptions for every building able to be found in Google Maps—so that more relevant information becomes integrated into and findable through textual parametric knowledge alone. 
However, this strategy would still fall short in queries about newer pieces of information, particularly those that arise \textit{after} a model has been trained. In such cases, real-time tool use remains indispensable for accurate and up-to-date query responses.

\paragraph{Application.} 
Achieving strong performance on this benchmark corresponds to benefits across several application domains in search, from information retrieval to product and people searches—tasks that can otherwise demand considerable manual effort. By reducing the labor involved, systems here stand to significantly improve user efficiency in these domains. 
Ostensibly, \textit{search} remains one of the many capabilities desired by people in the broader concept of a ``general AI assistant,'' which still remains loosely defined \cite{mialon2024gaia} in current literature.
This research constitutes an important step toward building AI systems and capabilities that address practical, real-world needs while also laying the groundwork for future enhancements in areas users \textit{genuinely} care about in their daily lives.

\section{Conclusion}

We presented \textsc{\textbf{B}rowsing \textbf{L}ost \textbf{U}nformed \textbf{R}ecollections} (BLUR), a demanding 573-question tip-of-the-tongue known-item search and reasoning benchmark that is both multilingual and multimodal on input for general AI assistants. Evaluations show that current LLMs, LLM-based agents, and general AI systems still fall well short of human-level performance on this task. We hope that our release will foster more human-centered research and drive improvements in a search challenge that resonates across multiple domains.

\section*{Limitations}

\paragraph{Answer Longevity.} As the internet is constantly changing, the sources used to answer these questions may not always remain accessible---webpages can be taken down or become unavailable. To address this, we recorded the ``sources of truth'' (i.e., the websites visited) that annotators relied on when creating validation chains for each question. As part of this data release, we periodically check that these, or alternate sources, remain available. If any links stop working, we shall remove the affected questions to maintain valid evaluations. Even in the event of widespread unavailability, it remains straightforward to create a new dataset measuring the same capabilities by following our established annotation procedures.

\paragraph{System Stochasticity.} This benchmark primarily focuses on evaluating \textit{capability}, without a need to specify or restrict to any sets of tools in requirement. However, as the underlying tools---often outside the control of individual system designers---in these systems can change and evolve, system capabilities may shift over time as well. For example, we originally sought to benchmark MagneticOne \cite{fourney2024magentic}—an agentic system powered by the AutoGen framework—but it lost its OCR capabilities during testing after the OpenAI API stopped processing OCR requests.\footnote{https://github.com/microsoft/autogen/issues/4482}
These issues highlight the difficulty in system evaluation, especially when certain and necessary system components fall outside the direct control of the system's designers.

\paragraph{Future Work.} 
A key future challenge lies in \textit{evaluating reasoning chains}, particularly when models arrive at the \textit{wrong answer}. While correctness is straightforward to confirm for successful outcomes, gauging how close a model was to the correct solution is a more nuanced question, especially apparent in tool-using reasoning agents \cite{prasad-etal-2023-receval}. Future research that develops methods to quantify the \textit{distance} between a model’s chain of thought and the correct reasoning process could provide more precise indicators for improving agent capabilities.
A second future challenge in evaluation lies in \textit{efficiency}---namely, the time required for a model to arrive at the correct answer, which is also of practical importance to end users. Evaluating efficiency becomes especially complex when the results from tool use are stochastic. Future research could focus on reliably measuring this aspect, perhaps by tracking token usage or other metrics that gauge the time and computational cost of each reasoning step. This would enable more informed comparisons of models’ performance in real-world scenarios, in addition to base comparisons of answer correctness.
Finally, a third challenge lies in \textit{multi-turn} evaluation. For queries that \textit{cannot} be answered in a single round, an exciting open question remains on how to best facilitate and evaluate answer-seeking strategies across multiple conversation rounds. It surmises that users may desire and benefit from collaborating with an assistant to navigate through incomplete information, refine their questions, and progressively build toward the correct answer---unlike traditional search which generally relies on well-defined queries---suggesting related ideas along the way that might help trigger recognition. 

\section*{Ethical Considerations}

Following data collection, we reviewed all data items manually and ensured that no Personally Identifiable Information (PII) was present in the data. Annotators were paid for their time at a rate of \$20 an hour. 

\section*{Acknowledgments}

We thank Chris McConnell, Annie Franco, Thomas Sciacolm, Selvan Sunitha Ravi, Devanshu Bansal, Matthew Toles, Douwe Kiela, Stephen Roller, Victor Sanh, \mbox{Aochong} Li, and Anubhav Jangra for helpful conversations during the completion of this paper.

\bibliography{custom}

\begin{thebibliography}{47}
\expandafter\ifx\csname natexlab\endcsname\relax\def\natexlab#1{#1}\fi

\bibitem[{Anthropic(2024)}]{claude_computer_use}
Anthropic. 2024.
\newblock Model card addendum: Claude 3.5 haiku and upgraded claude 3.5 sonnet.

\bibitem[{Arguello et~al.(2021)Arguello, Ferguson, Fine, Mitra, Zamani, and Diaz}]{arguello2021tip}
Jaime Arguello, Adam Ferguson, Emery Fine, Bhaskar Mitra, Hamed Zamani, and Fernando Diaz. 2021.
\newblock Tip of the tongue known-item retrieval: A case study in movie identification.
\newblock In \emph{Proceedings of the 2021 Conference on Human Information Interaction and Retrieval}, pages 5--14.

\bibitem[{Bhargav et~al.(2023)Bhargav, Schuth, and Hauff}]{bhargav2023music}
Samarth Bhargav, Anne Schuth, and Claudia Hauff. 2023.
\newblock When the music stops: Tip-of-the-tongue retrieval for music.
\newblock In \emph{Proceedings of the 46th International ACM SIGIR Conference on Research and Development in Information Retrieval}, pages 2506--2510.

\bibitem[{Bhargav et~al.(2022)Bhargav, Sidiropoulos, and Kanoulas}]{bhargav2022s}
Samarth Bhargav, Georgios Sidiropoulos, and Evangelos Kanoulas. 2022.
\newblock 'it's on the tip of my tongue' a new dataset for known-item retrieval.
\newblock In \emph{Proceedings of the Fifteenth ACM International Conference on Web Search and Data Mining}, pages 48--56.

\bibitem[{Bogers et~al.(2024)Bogers, G{\"a}de, Koolen, Petras, and Skov}]{bogers2024understanding}
Toine Bogers, Maria G{\"a}de, Marijn Koolen, Vivien Petras, and Mette Skov. 2024.
\newblock Understanding complex casual leisure information needs: an analysis of search requests for books, games, movies and music.
\newblock \emph{Journal of Documentation}.

\bibitem[{Borges et~al.(2024)Borges, Jha, Callan, and Martins}]{borges2024generalizable}
Lu{\'\i}s Borges, Rohan Jha, Jamie Callan, and Bruno Martins. 2024.
\newblock Generalizable tip-of-the-tongue retrieval with llm re-ranking.
\newblock In \emph{Proceedings of the 47th International ACM SIGIR Conference on Research and Development in Information Retrieval}, pages 2437--2441.

\bibitem[{Bowman and Dahl(2021)}]{bowman-dahl-2021-will}
Samuel~R. Bowman and George Dahl. 2021.
\newblock What will it take to fix benchmarking in natural language understanding?
\newblock In \emph{Proceedings of the 2021 Conference of the North American Chapter of the Association for Computational Linguistics: Human Language Technologies}, pages 4843--4855, Online. Association for Computational Linguistics.

\bibitem[{Chi et~al.(2024)Chi, Lin, Lin, and Klein}]{Chi2024CLARINETAL}
Yizhou Chi, Jessy Lin, Kevin Lin, and Dan Klein. 2024.
\newblock Clarinet: Augmenting language models to ask clarification questions for retrieval.
\newblock \emph{ArXiv}, abs/2405.15784.

\bibitem[{De~Vries et~al.(2020)De~Vries, Bahdanau, and Manning}]{de2020towards}
Harm De~Vries, Dzmitry Bahdanau, and Christopher Manning. 2020.
\newblock Towards ecologically valid research on language user interfaces.
\newblock \emph{arXiv preprint arXiv:2007.14435}.

\bibitem[{Deng et~al.(2023)Deng, Gu, Zheng, Chen, Stevens, Wang, Sun, and Su}]{NEURIPS2023_mind2web}
Xiang Deng, Yu~Gu, Boyuan Zheng, Shijie Chen, Sam Stevens, Boshi Wang, Huan Sun, and Yu~Su. 2023.
\newblock Mind2web: Towards a generalist agent for the web.
\newblock In \emph{Advances in Neural Information Processing Systems}, volume~36, pages 28091--28114. Curran Associates, Inc.

\bibitem[{Fourney et~al.(2024)Fourney, Bansal, Mozannar, Tan, Salinas, Niedtner, Proebsting, Bassman, Gerrits, Alber et~al.}]{fourney2024magentic}
Adam Fourney, Gagan Bansal, Hussein Mozannar, Cheng Tan, Eduardo Salinas, Friederike Niedtner, Grace Proebsting, Griffin Bassman, Jack Gerrits, Jacob Alber, et~al. 2024.
\newblock Magentic-one: A generalist multi-agent system for solving complex tasks.
\newblock \emph{arXiv preprint arXiv:2411.04468}.

\bibitem[{Fr{\"o}be et~al.(2023)Fr{\"o}be, Schmidt, and Hagen}]{frobe2023large}
Maik Fr{\"o}be, Eric~Oliver Schmidt, and Matthias Hagen. 2023.
\newblock A large-scale dataset for known-item question performance prediction.
\newblock In \emph{QPP++@ ECIR}, pages 13--19.

\bibitem[{Ge et~al.(2024)Ge, Hua, Mei, Tan, Xu, Li, Zhang et~al.}]{ge2024openagi}
Yingqiang Ge, Wenyue Hua, Kai Mei, Juntao Tan, Shuyuan Xu, Zelong Li, Yongfeng Zhang, et~al. 2024.
\newblock Openagi: When llm meets domain experts.
\newblock \emph{Advances in Neural Information Processing Systems}, 36.

\bibitem[{Grattafiori et~al.(2024)Grattafiori, Dubey, Jauhri, Pandey, Kadian, Al-Dahle, Letman, Mathur, Schelten, Vaughan et~al.}]{grattafiori2024llama}
Aaron Grattafiori, Abhimanyu Dubey, Abhinav Jauhri, Abhinav Pandey, Abhishek Kadian, Ahmad Al-Dahle, Aiesha Letman, Akhil Mathur, Alan Schelten, Alex Vaughan, et~al. 2024.
\newblock The llama 3 herd of models.
\newblock \emph{arXiv preprint arXiv:2407.21783}.

\bibitem[{Guo et~al.(2025)Guo, Yang, Zhang, Song, Zhang, Xu, Zhu, Ma, Wang, Bi et~al.}]{guo2025deepseek}
Daya Guo, Dejian Yang, Haowei Zhang, Junxiao Song, Ruoyu Zhang, Runxin Xu, Qihao Zhu, Shirong Ma, Peiyi Wang, Xiao Bi, et~al. 2025.
\newblock Deepseek-r1: Incentivizing reasoning capability in llms via reinforcement learning.
\newblock \emph{arXiv preprint arXiv:2501.12948}.

\bibitem[{Hendrycks et~al.(2021)Hendrycks, Burns, Basart, Zou, Mazeika, Song, and Steinhardt}]{hendrycks2021measuring}
Dan Hendrycks, Collin Burns, Steven Basart, Andy Zou, Mantas Mazeika, Dawn Song, and Jacob Steinhardt. 2021.
\newblock Measuring massive multitask language understanding.
\newblock In \emph{International Conference on Learning Representations}.

\bibitem[{Hu et~al.(2024)Hu, Zhao, Wei, Chai, Ma, Wang, Wang, Su, Xu, Zhu, Cheng, Yuan, Li, Kuang, Yang, Yang, and Wu}]{hu2024infiagent}
Xueyu Hu, Ziyu Zhao, Shuang Wei, Ziwei Chai, Qianli Ma, Guoyin Wang, Xuwu Wang, Jing Su, Jingjing Xu, Ming Zhu, Yao Cheng, Jianbo Yuan, Jiwei Li, Kun Kuang, Yang Yang, Hongxia Yang, and Fei Wu. 2024.
\newblock {I}nfi{A}gent-{DAB}ench: Evaluating agents on data analysis tasks.
\newblock In \emph{Proceedings of the 41st International Conference on Machine Learning}, volume 235 of \emph{Proceedings of Machine Learning Research}, pages 19544--19572. PMLR.

\bibitem[{{HuggingFace}(2024)}]{hfagents}
{HuggingFace}. 2024.
\newblock License to call: Introducing transformers agents 2.0.
\newblock \url{https://huggingface.co/blog/agents}.

\bibitem[{Jimenez et~al.(2024)Jimenez, Yang, Wettig, Yao, Pei, Press, and Narasimhan}]{jimenez2024swebench}
Carlos~E Jimenez, John Yang, Alexander Wettig, Shunyu Yao, Kexin Pei, Ofir Press, and Karthik~R Narasimhan. 2024.
\newblock {SWE}-bench: Can language models resolve real-world github issues?
\newblock In \emph{The Twelfth International Conference on Learning Representations}.

\bibitem[{Kiela et~al.(2021)Kiela, Bartolo, Nie, Kaushik, Geiger, Wu, Vidgen, Prasad, Singh, Ringshia, Ma, Thrush, Riedel, Waseem, Stenetorp, Jia, Bansal, Potts, and Williams}]{kiela-etal-2021-dynabench}
Douwe Kiela, Max Bartolo, Yixin Nie, Divyansh Kaushik, Atticus Geiger, Zhengxuan Wu, Bertie Vidgen, Grusha Prasad, Amanpreet Singh, Pratik Ringshia, Zhiyi Ma, Tristan Thrush, Sebastian Riedel, Zeerak Waseem, Pontus Stenetorp, Robin Jia, Mohit Bansal, Christopher Potts, and Adina Williams. 2021.
\newblock Dynabench: Rethinking benchmarking in {NLP}.
\newblock In \emph{Proceedings of the 2021 Conference of the North American Chapter of the Association for Computational Linguistics: Human Language Technologies}, pages 4110--4124, Online. Association for Computational Linguistics.

\bibitem[{Kiela et~al.(2023)Kiela, Thrush, Ethayarajh, and Singh}]{kiela2023plottingprogress}
Douwe Kiela, Tristan Thrush, Kawin Ethayarajh, and Amanpreet Singh. 2023.
\newblock Plotting progress in ai.
\newblock \emph{Contextual AI Blog}.

\bibitem[{Kim et~al.(2024)Kim, Stora{\"\i}, and Hwang}]{kim-etal-2024-intended}
Jongho Kim, Romain Stora{\"\i}, and Seung-won Hwang. 2024.
\newblock Intended target identification for anomia patients with gradient-based selective augmentation.
\newblock In \emph{Findings of the Association for Computational Linguistics: EMNLP 2024}, pages 10513--10527, Miami, Florida, USA. Association for Computational Linguistics.

\bibitem[{Li et~al.(2023)Li, Zhao, Yu, Song, Li, Yu, Li, Huang, and Li}]{li-etal-2023-api}
Minghao Li, Yingxiu Zhao, Bowen Yu, Feifan Song, Hangyu Li, Haiyang Yu, Zhoujun Li, Fei Huang, and Yongbin Li. 2023.
\newblock {API}-bank: A comprehensive benchmark for tool-augmented {LLM}s.
\newblock In \emph{Proceedings of the 2023 Conference on Empirical Methods in Natural Language Processing}, pages 3102--3116, Singapore. Association for Computational Linguistics.

\bibitem[{Liao and Xiao(2023)}]{liao2023rethinking}
Q~Vera Liao and Ziang Xiao. 2023.
\newblock Rethinking model evaluation as narrowing the socio-technical gap.
\newblock \emph{arXiv preprint arXiv:2306.03100}.

\bibitem[{Lin et~al.(2023)Lin, Lo, Gonzalez, and Klein}]{lin-etal-2023-decomposing}
Kevin Lin, Kyle Lo, Joseph Gonzalez, and Dan Klein. 2023.
\newblock Decomposing complex queries for tip-of-the-tongue retrieval.
\newblock In \emph{Findings of the Association for Computational Linguistics: EMNLP 2023}, pages 5521--5533, Singapore. Association for Computational Linguistics.

\bibitem[{Liu et~al.(2024{\natexlab{a}})Liu, Yu, Zhang, Xu, Lei, Lai, Gu, Ding, Men, Yang, Zhang, Deng, Zeng, Du, Zhang, Shen, Zhang, Su, Sun, Huang, Dong, and Tang}]{liu2024agentbench}
Xiao Liu, Hao Yu, Hanchen Zhang, Yifan Xu, Xuanyu Lei, Hanyu Lai, Yu~Gu, Hangliang Ding, Kaiwen Men, Kejuan Yang, Shudan Zhang, Xiang Deng, Aohan Zeng, Zhengxiao Du, Chenhui Zhang, Sheng Shen, Tianjun Zhang, Yu~Su, Huan Sun, Minlie Huang, Yuxiao Dong, and Jie Tang. 2024{\natexlab{a}}.
\newblock \href {https://openreview.net/forum?id=zAdUB0aCTQ} {Agentbench: Evaluating {LLM}s as agents}.
\newblock In \emph{The Twelfth International Conference on Learning Representations}.

\bibitem[{Liu et~al.(2024{\natexlab{b}})Liu, Blodgett, Cheung, Liao, Olteanu, and Xiao}]{liu-etal-2024-ecbd}
Yu~Lu Liu, Su~Lin Blodgett, Jackie Cheung, Q.~Vera Liao, Alexandra Olteanu, and Ziang Xiao. 2024{\natexlab{b}}.
\newblock {ECBD}: Evidence-centered benchmark design for {NLP}.
\newblock In \emph{Proceedings of the 62nd Annual Meeting of the Association for Computational Linguistics (Volume 1: Long Papers)}, pages 16349--16365, Bangkok, Thailand. Association for Computational Linguistics.

\bibitem[{Meier et~al.(2021)Meier, Bogers, G{\"a}de, and Ebdrup~Thomsen}]{meier2021towards}
Florian Meier, Toine Bogers, Maria G{\"a}de, and Line Ebdrup~Thomsen. 2021.
\newblock Towards understanding complex known-item requests on reddit.
\newblock In \emph{Proceedings of the 32nd ACM Conference on Hypertext and Social Media}, pages 143--154.

\bibitem[{Mialon et~al.(2024)Mialon, Fourrier, Wolf, LeCun, and Scialom}]{mialon2024gaia}
Gr{\'e}goire Mialon, Cl{\'e}mentine Fourrier, Thomas Wolf, Yann LeCun, and Thomas Scialom. 2024.
\newblock {GAIA}: a benchmark for general {AI} assistants.
\newblock In \emph{The Twelfth International Conference on Learning Representations}.

\bibitem[{{Mistral AI Team}(2024)}]{mistrallechat}
{Mistral AI Team}. 2024.
\newblock Mistral has entered the chat.
\newblock \url{https://mistral.ai/news/mistral-chat/}.

\bibitem[{Mun et~al.(2024)Mun, Jiang, Liang, Cheong, DeCario, Choi, Kohno, and Sap}]{mun2024participAI}
Jimin Mun, Liwei Jiang, Jenny Liang, Inyoung Cheong, Nicole DeCario, Yejin Choi, Tadayoshi Kohno, and Maarten Sap. 2024.
\newblock Particip-ai: A democratic surveying framework for anticipating future ai use cases, harms and benefits.
\newblock In \emph{AIES}.

\bibitem[{Nguyen et~al.(2024)Nguyen, Lai, Yoon, Rossi, Zhao, Zhang, Mathur, Lipka, Wang, Bui et~al.}]{nguyen2024dynasaur}
Dang Nguyen, Viet~Dac Lai, Seunghyun Yoon, Ryan~A Rossi, Handong Zhao, Ruiyi Zhang, Puneet Mathur, Nedim Lipka, Yu~Wang, Trung Bui, et~al. 2024.
\newblock Dynasaur: Large language agents beyond predefined actions.
\newblock \emph{arXiv preprint arXiv:2411.01747}.

\bibitem[{OpenAI(2024{\natexlab{a}})}]{gpt4o}
OpenAI. 2024{\natexlab{a}}.
\newblock Gpt-4o system card.

\bibitem[{OpenAI(2024{\natexlab{b}})}]{gpto1}
OpenAI. 2024{\natexlab{b}}.
\newblock Openai o1 system card.

\bibitem[{OpenAI(2025)}]{cua2025}
OpenAI. 2025.
\newblock Computer-using agent: Introducing a universal interface for ai to interact with the digital world.

\bibitem[{Patil et~al.(2024)Patil, Zhang, Wang, and Gonzalez}]{patil2023gorilla}
Shishir~G Patil, Tianjun Zhang, Xin Wang, and Joseph~E. Gonzalez. 2024.
\newblock Gorilla: Large language model connected with massive {API}s.
\newblock In \emph{The Thirty-eighth Annual Conference on Neural Information Processing Systems}.

\bibitem[{{Perplexity Team}(2024)}]{perplexitypro}
{Perplexity Team}. 2024.
\newblock Pro search: Upgraded for more advanced problem-solving.
\newblock \url{https://www.perplexity.ai/}.

\bibitem[{Petroni et~al.(2021)Petroni, Piktus, Fan, Lewis, Yazdani, De~Cao, Thorne, Jernite, Karpukhin, Maillard, Plachouras, Rockt{\"a}schel, and Riedel}]{petroni-etal-2021-kilt}
Fabio Petroni, Aleksandra Piktus, Angela Fan, Patrick Lewis, Majid Yazdani, Nicola De~Cao, James Thorne, Yacine Jernite, Vladimir Karpukhin, Jean Maillard, Vassilis Plachouras, Tim Rockt{\"a}schel, and Sebastian Riedel. 2021.
\newblock {KILT}: a benchmark for knowledge intensive language tasks.
\newblock In \emph{Proceedings of the 2021 Conference of the North American Chapter of the Association for Computational Linguistics: Human Language Technologies}, pages 2523--2544, Online. Association for Computational Linguistics.

\bibitem[{Phang et~al.(2022)Phang, Chen, Huang, and Bowman}]{phang-etal-2022-adversarially}
Jason Phang, Angelica Chen, William Huang, and Samuel~R. Bowman. 2022.
\newblock Adversarially constructed evaluation sets are more challenging, but may not be fair.
\newblock In \emph{Proceedings of the First Workshop on Dynamic Adversarial Data Collection}, pages 62--62, Seattle, WA. Association for Computational Linguistics.

\bibitem[{Prasad et~al.(2023)Prasad, Saha, Zhou, and Bansal}]{prasad-etal-2023-receval}
Archiki Prasad, Swarnadeep Saha, Xiang Zhou, and Mohit Bansal. 2023.
\newblock {R}e{CE}val: Evaluating reasoning chains via correctness and informativeness.
\newblock In \emph{Proceedings of the 2023 Conference on Empirical Methods in Natural Language Processing}, pages 10066--10086, Singapore. Association for Computational Linguistics.

\bibitem[{Rein et~al.(2024)Rein, Hou, Stickland, Petty, Pang, Dirani, Michael, and Bowman}]{rein2024gpqa}
David Rein, Betty~Li Hou, Asa~Cooper Stickland, Jackson Petty, Richard~Yuanzhe Pang, Julien Dirani, Julian Michael, and Samuel~R. Bowman. 2024.
\newblock {GPQA}: A graduate-level google-proof q\&a benchmark.
\newblock In \emph{First Conference on Language Modeling}.

\bibitem[{Sainz et~al.(2023)Sainz, Campos, Garc{\'\i}a-Ferrero, Etxaniz, de~Lacalle, and Agirre}]{sainz-etal-2023-nlp}
Oscar Sainz, Jon Campos, Iker Garc{\'\i}a-Ferrero, Julen Etxaniz, Oier~Lopez de~Lacalle, and Eneko Agirre. 2023.
\newblock {NLP} evaluation in trouble: On the need to measure {LLM} data contamination for each benchmark.
\newblock In \emph{Findings of the Association for Computational Linguistics: EMNLP 2023}, pages 10776--10787, Singapore. Association for Computational Linguistics.

\bibitem[{Wagstaff(2012)}]{wagstaff2012machine}
Kiri~L. Wagstaff. 2012.
\newblock Machine learning that matters.
\newblock In \emph{Proceedings of the 29th International Conference on Machine Learning, 2012}, pages 529--536.

\bibitem[{Wei et~al.(2024)Wei, Karina, Chung, Jiao, Papay, Glaese, Schulman, and Fedus}]{wei2024measuring}
Jason Wei, Nguyen Karina, Hyung~Won Chung, Yunxin~Joy Jiao, Spencer Papay, Amelia Glaese, John Schulman, and William Fedus. 2024.
\newblock Measuring short-form factuality in large language models.
\newblock \emph{arXiv preprint arXiv:2411.04368}.

\bibitem[{Xie et~al.(2024)Xie, Zhang, Chen, Li, Zhao, Cao, Toh, Cheng, Shin, Lei et~al.}]{OSWorld}
Tianbao Xie, Danyang Zhang, Jixuan Chen, Xiaochuan Li, Siheng Zhao, Ruisheng Cao, Jing~Hua Toh, Zhoujun Cheng, Dongchan Shin, Fangyu Lei, et~al. 2024.
\newblock Osworld: Benchmarking multimodal agents for open-ended tasks in real computer environments.
\newblock \emph{Advances in Neural Information Processing Systems}, 37:52040--52094.

\bibitem[{Yoran et~al.(2024)Yoran, Amouyal, Malaviya, Bogin, Press, and Berant}]{yoran2024assistantbench}
Ori Yoran, Samuel~Joseph Amouyal, Chaitanya Malaviya, Ben Bogin, Ofir Press, and Jonathan Berant. 2024.
\newblock {A}ssistant{B}ench: Can web agents solve realistic and time-consuming tasks?
\newblock In \emph{Proceedings of the 2024 Conference on Empirical Methods in Natural Language Processing}, pages 8938--8968, Miami, Florida, USA. Association for Computational Linguistics.

\bibitem[{Zheng et~al.(2023)Zheng, Chiang, Sheng, Zhuang, Wu, Zhuang, Lin, Li, Li, Xing, Zhang, Gonzalez, and Stoica}]{zheng2023judging}
Lianmin Zheng, Wei-Lin Chiang, Ying Sheng, Siyuan Zhuang, Zhanghao Wu, Yonghao Zhuang, Zi~Lin, Zhuohan Li, Dacheng Li, Eric Xing, Hao Zhang, Joseph~E. Gonzalez, and Ion Stoica. 2023.
\newblock Judging {LLM}-as-a-judge with {MT}-bench and chatbot arena.
\newblock In \emph{Thirty-seventh Conference on Neural Information Processing Systems Datasets and Benchmarks Track}.

\end{thebibliography}
\bibliographystyle{acl_natbib}

\appendix

\section{Annotation Details}
\label{sec:annotation_details}

\paragraph{Annotators.} A total of 12 annotators participated in the data collection process. Age-wise, 10 (83\%) annotators were within the 26–35 age range, and 2 (17\%) were within the 18–25 age range. 7 (58\%) identify as women, 5 (42\%) identify as men. 11 (92\%) hold a Bachelor's degree, and 1 (8\%) holds a Master's degree. 

\paragraph{Query Creation Guidelines.} Have you ever been caught at a loss for words? Where you know something exists—and can describe it—but don’t know the exact or key phrase to search for on Google? We want to augment the dataset of the provided questions below. We’re looking for creativity in this process, not for only variations of what we already have. Try to be as diverse as you can in the examples you come up with!

When creating examples, it’s helpful to recall a time in the past when you tried to remember the name of something—and knew how it looked, how it felt, how it sounded, etc.—but just couldn’t recall its name. What did it feel like when you were trying to recall what it was called?

The following list is provided for inspiration of potential domains to choose from when creating your dataset. They are by no means exhaustive—do feel free (in fact, it’s very encouraged!) to come up with examples that fall outside of these domains and descriptions. By no means do they need to be famous or well-known. In fact, it’s usually the obscure ones that are hard to remember! 

\begin{itemize}
    \item Places—i.e. buildings, places you’ve been to, and more.
    \item People
    \item Science—i.e. academic papers.
    \item Arts \& Culture—i.e. artwork, artisan crafts, books and novels, etc.
    \item Entertainment \& Media—i.e. music, TV shows, movie characters, commercials, animations, etc. 
    \item Technology \& Tools—i.e. apps, old gadgets, etc.
    \item Experiences—i.e. rituals, local adventures, etc.
    \item Programming—i.e. specific github repos
    \item Websites
    \item Others—feel free (\& it’s encouraged!) to come up with your own!
\end{itemize}

No restrictions are placed on the language of the information present in the query you have, but please make sure that the instructions of the query are in English. If you find it hard to describe, in natural language, a piece of information that you remember about the entity in question, feel free to include it as part of a file upload! Please structure your prompt in how you might seek help online or from a friend.

When you’ve finished creating your prompt, please go ahead and spend some time trying to find the item in question! When doing so, please record the steps you took to try to find the answer to the query (even if they were wrong!). Once you’ve found the item, record it down in the form of a short string. Some general guidelines:

\begin{itemize}
    \item Make sure the answer to your question is unambiguous. 
    \item Make sure your question can be answered in a reasonable amount of time by a human annotator.
    \item Make sure the answer to your question is at most a few words.
    \item Make sure the answer to your question does not exist on the internet in plain text.
\end{itemize}

\textit{Here, one of the example questions shown to annotators is provided; others are left out for privacy reasons.}

\begin{displayquote}
\raggedright
    
    \textit{Example Question 1.} I remember this song. It's in Chinese. The start of this song is like the piano opening of the song I’ve attached, though I'm pretty sure the start of the song I'm thinking of was not in piano. It's by someone pretty famous, and I remember going to their concert and hearing this song, I think, sometime around 2008-2012, in Beijing, China. A lyric I remember from the song is like ``\chinese{我不想再写}'' and ``\chinese{满天的风雪}'' something. The concert was pretty pink, and everyone had pink lightsticks. It was like a sea of pink. There was something about broken butterflies in the song. I think the song's theme was about breaking up or love. What is the name of the song I'm thinking of? \textbf{File Input:} MP3 file to Yuri On Ice.
    
\end{displayquote}

\paragraph{Query Validation Guidelines.} Please identify the entity mentioned in the following query. You may consult any files or programs you wish, and you are especially encouraged to use a web browser for assistance. Note that this query was originally asked by a user on December 15, 2024. 

Your response should be a concise string. Once you have formulated your answer, record it and then click Confirm. Afterward, compare your answer to the indicated answer. If the two do not match, please provide a brief explanation—either affirming why your answer is correct or explaining any mistake you may have found in your reasoning.

Finally, state whether you believe the query is unambiguous, specifically addressing whether the indicated answer is the only valid option you discovered during your process of arriving at a solution.

\section{Prompts}
\label{sec:prompts}

\paragraph{LLM judge.} You are a judge. You will be provided with three pieces of information:

\begin{itemize}
    \item \textbf{Question}: A question asked by the user.
    \item \textbf{Ground Truth Answer}: The correct, specific answer to that question.
    \item \textbf{User-Provided Answer}: A potentially correct or incorrect attempt at answering the question.
\end{itemize}

Your task is to determine if the User-Provided Answer unambiguously corresponds to the same information/entity described by the Ground Truth Answer, in the context of the Question.

\begin{itemize}
    \item If the User-Provided Answer clearly and uniquely identifies the same information as the Ground Truth Answer (even if phrased differently or missing non-essential details), then it should be considered \textbf{correct}.
    \item If it is ambiguous, incomplete, or could refer to multiple different answers, then it must be considered \textbf{incorrect}.
\end{itemize}

Please output your reasoning, alongside one of the following: ``Correct'' if the User-Provided Answer unambiguously matches the Ground Truth Answer, and ``Incorrect'' otherwise.

\section{Reasoning Traces}
\label{sec:reasoning_traces}

Due to the lack of transparency in reasoning traces in o1, here, we use DeepSeek r1 to show examples of varying behavior in reasoning traces in cases without tool use. In the first instance below, the model is capable of recalling the correct book a character is in when directly prompted with a vague description of the story, highlighting how non-tool-use models can be effective at piecing together pieces of information already contained in parametric knowledge. 

\begin{displayquote}
\raggedright

Okay, let's try to figure out the name of the main female character the user is talking about. The novel is about a couple who have to go through dating and marriage again because the woman loses her memory after a car accident less than a year into their marriage. It's based on a true story and focuses on themes of love, commitment, and the impact of memory loss.

\textbf{First, I remember there's a novel called "The Vow" which is based on a true story.} The real-life couple is Kim and Krickitt Carpenter, who experienced a similar situation where Krickitt had amnesia after a car accident and couldn't remember her marriage...

\end{displayquote}

When the entity may not be in parametric knowledge, below, DeepSeek r1 appears to make an educated guess.

\begin{displayquote}
\raggedright

...Since the user's query is from December 2024, this might be the upcoming content they saw a preview of. If Season 8 was releasing episodes around that time (webpages 1 and 2 have episodes 4 and 5 in December 2024 and January 2025), \textbf{maybe the character Oluwabunmi was introduced in Season 8.}...

\end{displayquote}

For systems with tool-use capabilities, here, ChatGPT 4o with Deep Research misreads the text in an image ``Kapital Kitchen'' as ``Krispy Chicken,'' which orchestrates a series of wrong searches. A cropped version of the image centering on the mistake is shown for display purposes.

\begin{displayquote}
\raggedright

\includegraphics[width=0.98\linewidth]{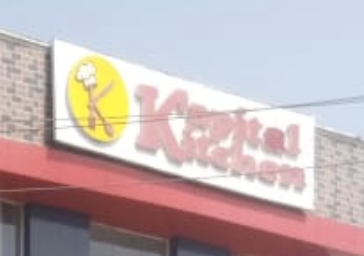}

Thanks for sharing the image! I see the restaurant in the photo, which is "Krispy Chicken."...

...Identifying Lounge Near Krispy Chicken...

...I'm mapping out a lounge near Krispy Chicken...

\end{displayquote}

In a prompt that specifies a travel site ``launched in the early 2000s,'' Operator begins and continues its search for early 2000s travel sites that fit its description. Despite this queried constraint, it narrows in on Expedia, which was launched in 1996---even when it continuously encounters said information during its search.  

\begin{displayquote}
\raggedright

Searching for flight and hotel tracker

...

Scrolling for early 2000s website info

...

Refining search for early 2000s travel sites

...

Scrolling for early 2000s travel sites

...

Researching Expedia's design and features

\includegraphics[width=0.98\linewidth]{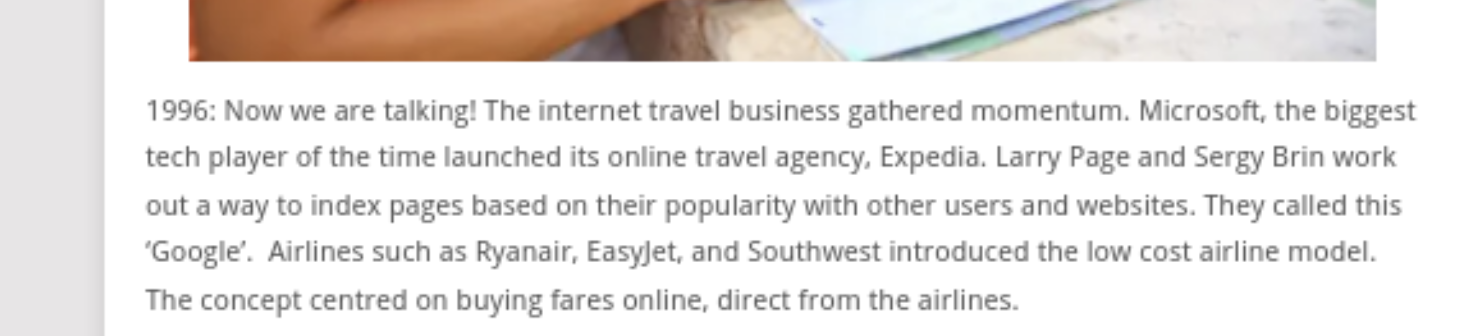}

Exploring Web Design Museum for insights

\includegraphics[width=0.98\linewidth]{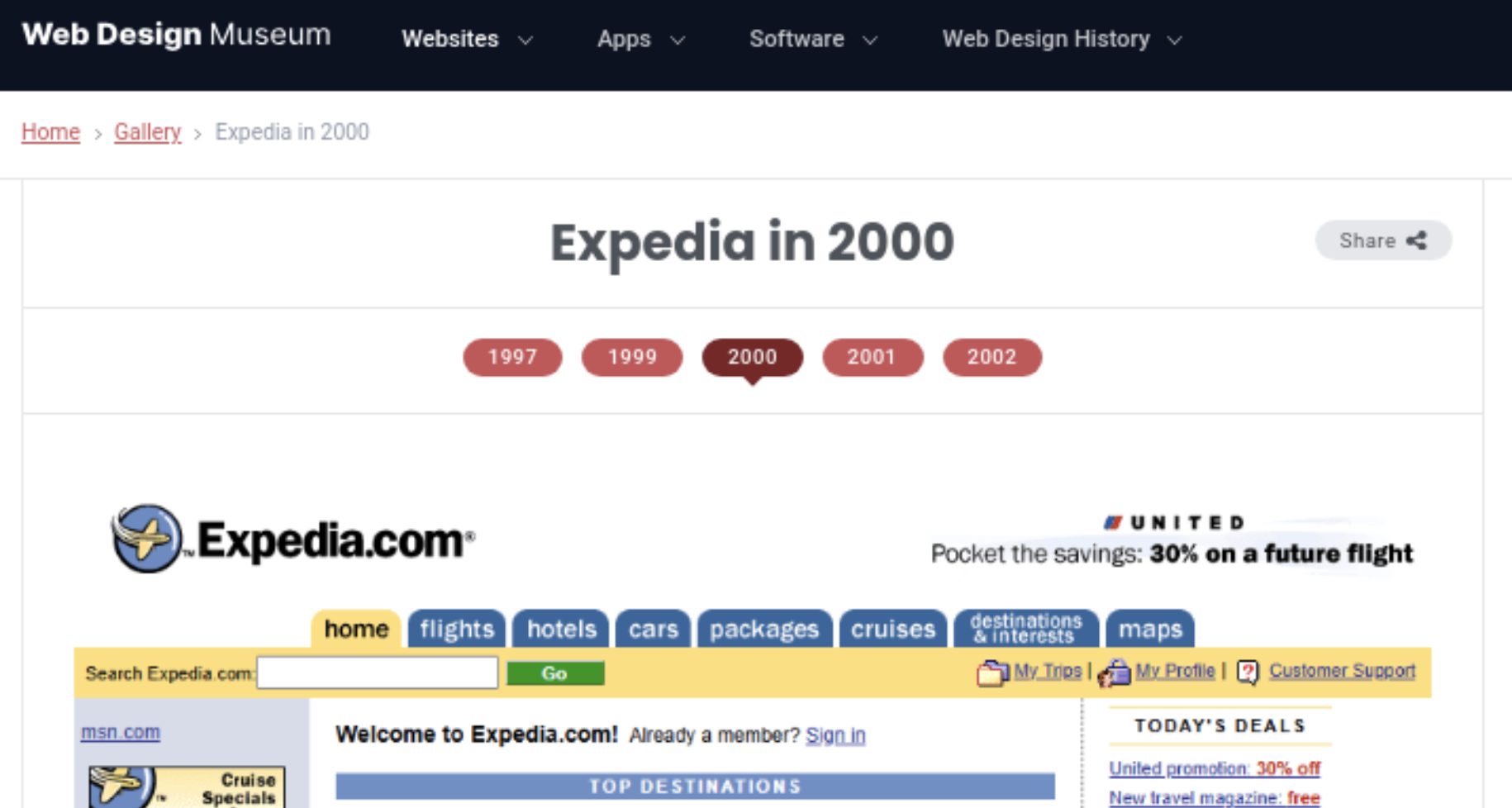}

Identifying Expedia as user’s reference

\textbf{Final Answer: Expedia, launched in 1996.}

\end{displayquote}

\end{document}